# The Bayesian Decision Tree Technique with a Sweeping Strategy

Vitaly SCHETININ, Jonathan E. FIELDSEND, *Member, IEEE*, Derek PARTRIDGE, Wojtek J. KRZANOWSKI, Richard M. EVERSON, Trevor C. BAILEY, and Adolfo HERNANDEZ

*Abstract*--**The uncertainty of classification outcomes is of crucial importance for many safety critical applications including, for example, medical diagnostics. In such applications the uncertainty of classification can be reliably estimated within a Bayesian model averaging technique that allows the use of prior information. Decision Tree (DT) classification models used within such a technique gives experts additional information by making this classification scheme observable. The use of the Markov Chain Monte Carlo (MCMC) methodology of stochastic sampling makes the Bayesian DT technique feasible to perform. However, in practice, the MCMC technique may become stuck in a particular DT which is far away from a region with a maximal posterior. Sampling such DTs causes bias in the posterior estimates, and as a result the evaluation of classification uncertainty may be incorrect. In a particular case, the negative effect of such sampling may be reduced by giving additional prior information on the shape of DTs. In this paper we describe a new approach based on sweeping the DTs without additional priors on the favorite shape of DTs. The performances of Bayesian DT techniques with the standard and sweeping strategies are compared on a synthetic data as well as on real datasets. Quantitatively evaluating the uncertainty in terms of entropy of class posterior probabilities, we found that the sweeping strategy is superior to the standard strategy.**

*Index Terms*--**Classification, Probability, Trees, Uncertainty.**

## I. INTRODUCTION

The uncertainty of classification outcomes is of crucial importance for many safety critical applications such as medical diagnostics and prediction of survival of patient after injuries. In such applications Bayesian model averaging provide reliable estimates of the classification uncertainty. The use of Decision Tree (DT) classification models within a Bayesian averaging framework gives experts additional information by making the classification scheme observable [1, 2].

The main idea of using DT classification models is to recursively partition data points in an axis-parallel manner. Such models provide natural feature selection and uncover the most important features for the classification. The resultant DT classification models are easily interpretable by users.

Generally, a DT is a hierarchical system consisting of splitting and terminal nodes. DTs are binary if the splitting nodes ask a specific question and then divide the data points into two disjoint subsets. The terminal node assigns all data points falling in that node to the class whose points are prevalent. Within a Bayesian framework, the class posterior distribution is calculated for each terminal node, which makes the Bayesian integration computationally expensive [1, 2].

Breiman *et al.* [1] provided the Bayesian generalization of tree models required to evaluate the posterior distribution. To make the Bayesian averaging DTs a feasible approach, Chipman *et al.* [3] have suggested the use of the Markov Chain Monte Carlo (MCMC) technique, making a stochastic sample from the posterior distribution.

Sampling across DT models of variable dimensionality, the above technique exploits a Reversible Jump (RJ) extension suggested by Green [4]. When prior information is not distorted and the number of samples is reasonably large, the RJ MCMC technique, making birth, death, change-question, and change-rule moves, explores the posterior distribution and as a result provides accurate estimates of the posterior. However, in practice the lack of prior information brings bias in the posterior estimates, and as a result the evaluation of classification uncertainty may be incorrect [5].

Within the RJ MCMC technique, the prior on the number of splitting nodes should be given properly. Otherwise, most samples may be taken from the posterior calculated for DTs that are located far away from a region containing the desired DT models. Likewise, when the prior on the number of splits is assigned to be uniform, the minimal number of data points allowed to be at nodes may be set inappropriately small. In this case, the DTs will grow excessively and most of the samples will be taken from the posterior distribution calculated for over-fitted DTs. As a result, the use of inappropriately assigned priors leads to poor results [6].

Clearly, the lack of prior knowledge on the favored DT structure, which often happens in practice, increases the uncertainty in results of the Bayesian averaging DTs. In this paper we aim to decrease the uncertainty of classification outcomes by using a new Bayesian strategy of sampling DT models. The main idea behind this strategy is to explicitly assign the prior probability of further splitting DT nodes

Manuscript received September 5, 2004. This work was supported by the EPSRC under Grant No. GR/R24357/01.
V. Schetinin, J.E. Fieldsend, D. Partridge, W.J. Krzanowski, R.M. Everson, T.C. Bailey, and A. Hernandez are with the School of Engineering, Computer Science and Mathematics, University of Exeter, EX4 4QF, UK (e-mails: {v.schetinin, j.e.fieldsend, d.partridge, w.j.krzanowski, r.m.everson, t.c.bailey, a.hernandez}@exeter.ac.uk).

dependent on the range of values within which the number of data points will be not less than a given number.

The classification uncertainty of the Bayesian DT techniques with the standard and sweeping strategies is evaluated in terms of entropy of class posterior probabilities as described in [7]. The evaluations are made on an artificial problem and some real datasets taken from the UCI Machine Learning Repository [8] as well as on real trauma dataset taken from the London Emergency Centre.

The remaining part of the paper is organized as follows. Sections II and III describe the standard and suggested Bayesian DT techniques. Then section IV describes the experiments conducted to compare the performance and the classification uncertainty of these techniques. Finally, section V concludes the paper.

## II. THE BAYESIAN DECISION TREE TECHNIQUE

### A. The Basis of Bayesian Model Averaging

In general, the predictive distribution we are interested in is written as an integral over parameters $\theta$ of the classification model

$$p(y \mid \mathbf{x}, \mathbf{D}) = \int_\theta p(y \mid \mathbf{x}, \theta, \mathbf{D}) p(\theta \mid \mathbf{D}) d\theta \qquad (1)$$

where $y$ is the predicted class $(1, \ldots, C)$, $\mathbf{x} = (x_1, \ldots, x_m)$ is the $m$-dimensional input vector, and $\mathbf{D}$ denotes the given training data.

The integral (1) can be analytically calculated only in simple cases. In practice, part of the integrand in (1), which is the posterior density of $\theta$ conditioned on the data $\mathbf{D}$, $p(\theta \mid \mathbf{D})$, cannot usually be evaluated. However if values $\theta^{(1)}, \ldots, \theta^{(N)}$ are drawn from the posterior distribution $p(\theta \mid \mathbf{D})$, we can write

$$p(y \mid \mathbf{x}, \mathbf{D}) \approx \sum_{i=1}^{N} p(y \mid \mathbf{x}, \theta^{(i)}, \mathbf{D}) p(\theta^{(i)} \mid \mathbf{D})$$
$$= \frac{1}{N} \sum_{i=1}^{N} p(y \mid \mathbf{x}, \theta^{(i)}, \mathbf{D}). \qquad (2)$$

This is the basis of the MCMC technique for approximating integrals [5]. To perform the approximation, we need to generate random samples from $p(\theta \mid \mathbf{D})$ by running a Markov Chain until it has converged to a stationary distribution. After this we can draw samples from this Markov Chain and estimate the predictive posterior density (2).

Let us now define a classification problem presented by data $(\mathbf{x}_i, y_i)$, $i = 1, \ldots, n$, where $n$ is the number of data points and $y_i \in \{1, \ldots, C\}$ is a categorical response. Using DTs for classification, we need to determine the probability $\varphi_{ij}$ with which a datum $\mathbf{x}$ is assigned by terminal node $i = 1, \ldots, k$ to the $j$th class, where $k$ is the number of terminal nodes in the DT. Initially we can assign a $(C - 1)$-dimensional Dirichlet prior for each terminal node so that $p(\varphi_i \mid \theta) = \text{Di}_{C-1}(\varphi_i \mid \alpha)$, where $\varphi_i$ = $(\varphi_{i1}, \ldots, \varphi_{iC})$, $\theta$ is the vector of DT parameters, and $\alpha = (\alpha_1, \ldots, \alpha_C)$ is a prior vector of constants given for all the classes.

The DT parameters are defined as $\theta = (s_i^{pos}, s_i^{var}, s_i^{rule})$, $i = 1, \ldots, k - 1$, where $s_i^{pos}$, $s_i^{var}$ and $s_i^{rule}$ define the *position*, *predictor* and *rule* of each splitting node, respectively. For these parameters the priors can be specified as follows. First we can define a maximal number of splitting nodes, say, $s_{max} = n - 1$, so $s_i^{pos} \in \{1, \ldots, s_{max}\}$. Second we draw any of the $m$ predictors from a uniform discrete distribution $U(1, \ldots, m)$ and assign $s_i^{var} \in \{1, \ldots, m\}$. Finally the candidate value for the splitting variable $x_j = s_i^{var}$ is drawn from a uniform discrete distribution $U(x_j^{(1)}, \ldots, x_j^{(N)})$, where $N$ is the total number of possible splitting rules for predictor $x_j$, either categorical or continuous.

Such priors allow the exploring of DTs which partition data in as many ways as possible, and therefore we can assume that each DT with the same number of terminal nodes is equally likely [5]. For this case the prior for a complete DT is described as follows

$$p(\theta) = \left\{\prod_{i=1}^{k-1} p(s_i^{rule} \mid s_i^{var}) p(s_i^{var})\right\} p(\{s_i^{pos}\}_1^{k-1}). \qquad (3)$$

Having set the priors on the parameters $\varphi$ and $\theta$, we can determine the marginal likelihood for the data given the classification tree. In the general case this likelihood can be written as a multinomial Dirichlet distribution [5]:

$$p(\mathbf{D} \mid \theta) = \left[\frac{\Gamma\{\alpha C\}}{\{\Gamma(\alpha)\}^C}\right]^k \prod_{i=1}^{C} \frac{\prod_{j}^{C} \Gamma(m_{ij} + \alpha_j)}{\Gamma(n_i + \sum_{j=1}^{C} \alpha_j)} \qquad (4)$$

where $n_i$ is the number of data points falling in the $i$th terminal node of which $m_{ij}$ points are of class $j$ and $\Gamma$ is a Gamma function.

### B. Sampling Large Decision Trees

To allow sampling DT models of variable dimensionality, the MCMC technique exploits the Reversible Jump extension [4]. This extension allows the MCMC technique to sample large DTs induced from real-world data. To implement the RJ MCMC technique Chipman *et al.* [3] and Denison *et al.* [5] have suggested exploring the posterior probability by using the following types of moves.

*Birth*. Randomly split the data points falling in one of the terminal nodes by a new splitting node with the variable and rule drawn from the corresponding priors.

*Death*. Randomly pick a splitting node with two terminal nodes and assign it to be one terminal with the united data points.

*Change-split*. Randomly pick a splitting node and assign it a new splitting variable and rule drawn from the corresponding priors.

*Change-rule*. Randomly pick a splitting node and assign it a new rule drawn from a given prior.

The first two moves, birth and death, are reversible and change the dimensionality of $\theta$ as described in [4]. The remaining moves provide jumps within the current dimensionality of $\theta$. Note that the change-split move is included to make "large" jumps which potentially increase the chance of sampling from a maximal posterior whilst the change-rule move does "local" jumps.

For the birth moves, the proposal ratio $R$ is written

$$R = \frac{q(\theta | \theta')p(\theta')}{q(\theta'| \theta)p(\theta)}, \quad (5)$$

where the $q(\theta | \theta')$ and $q(\theta'| \theta)$ are the proposed distributions, $\theta'$ and $\theta$ are $(k + 1)$ and $k$-dimensional vectors of DT parameters, respectively, and $p(\theta)$ and $p(\theta')$ are the probabilities of the DT with parameters $\theta$ and $\theta'$:

$$p(\theta) = \left\{\prod_{i=1}^{k-1}\frac{1}{N(s_i^{var})}\frac{1}{m}\right\}\frac{k}{S_k}\frac{1}{K}, \quad (6)$$

where $N(s_i^{var})$ is the total number of possible splitting rules for variable $s_i^{var}$, $S_k$ is the number of ways of constructing a DT with $k$ terminal nodes, and $K$ is the maximal number of terminal nodes, $K = n - 1$.

For binary DTs, as given from graph theory, the number $S_k$ is the *Catalan number*

$$S_k = \frac{1}{k+1}\binom{2k}{k}, \quad (7)$$

which for large $k$ becomes astronomically large, e.g., for $k \geq 25$, $S_k \geq (4.8)^{12}$.

The proposal distributions are as follows

$$q(\theta | \theta') = \frac{d_{k+1}}{D_{Q'}}, \quad (8)$$

$$q(\theta'| \theta) = \frac{b_k}{k}\frac{1}{N(s_k^{var})}\frac{1}{m}, \quad (9)$$

where $D_{Q1} = D_Q + 1$ is the number of splitting nodes whose branches are both terminal nodes.

Then the proposal ratio for a *birth* is given by

$$R = \frac{d_{k+1}}{b_k}\frac{k}{D_{Q1}}\frac{S_k}{S_{k+1}}. \quad (10)$$

The number $D_{Q1}$ in Eq. 10 is dependent on the DT structure and it is clear that $D_{Q1} < k \; \forall \; k = 1, …, K$. Analyzing Eq. 10, we can also assume $d_{k+1} = b_k$. Then letting the DTs grow, i.e., $k \to K$, and considering $S_{k+1} > S_k$, we can see that the value of $R \to c$, where $c$ is a constant lying between 0 and 1.

Alternatively, for the death moves the proposal ratio is written as

$$R = \frac{b_k}{d_{k-1}}\frac{D_Q}{(k-1)}\frac{S_k}{S_{k-1}}. \quad (11)$$

We can see that under the assumptions made for the birth moves, the right side in this equation is equal or more than 1, i.e., $R \geq 1$.

### C. The Use of Knowledge of the Favored Structure

For a case when there is knowledge of the favored structure of the DT, Chipman et al. [3] suggested a generalization of prior (3) by assuming that the prior probability of further split of the terminal nodes to be dependent on how many splits have already been made above them. For example, for the $i$th terminal node the probability of its splitting is written as

$$p_{split}(i) = \gamma(1 + d_i)^{-\delta}, \quad (12)$$

where $d_i$ is the number of splits made above $i$ and $\gamma, \delta \geq 0$ are given constants.

We can see that the larger $\delta$, the more the prior favors "bushy" trees. For $\delta = 0$ each DT with the same number of terminal nodes appears with the same prior probability.

### D. The Difficulties of Sampling Large Decision Trees

When DTs are induced from real-world data, the number of splitting nodes can be very large. In such cases the size of DTs can rationally decrease by defining a minimal number of data points, $p_{min}$, allowed to be in the splitting nodes [3, 5]. If the number of data points in new partitions made after the birth or change moves becomes less than a given number $p_{min}$, such moves are assigned unavailable, and the RJ MCMC algorithm resamples such moves.

When the moves are assigned unavailable, this distorts the proposal probabilities $p_b$, $p_d$, and $p_c$ given for the birth, death, and change moves, respectively. The larger the DT, the smaller the number of data points falling in the splitting nodes, and correspondingly the larger is the probability with which moves become unavailable. Resampling the unavailable moves makes the balance between the proposal probabilities biased.

To show that the balance of proposal probabilities can be biased, let us assume an example with probabilities $p_b$, $p_d$, and $p_c$ set equal to 0.2, 0.2, and 0.6, respectively, note that $p_b + p_d + p_c = 1$. Let the DTs be large so that the birth and change moves are assigned unavailable with probabilities $p_{bu}$ and $p_{cu}$ equal to 0.1 and 0.3, respectively. As a result, the birth and change moves are made with probabilities equal to $(p_b - p_{bu})$ and $(p_c - p_{cu})$, respectively. Let now emulate 10000 moves with the given proposal probabilities. The resultant probabilities are shown in Fig. 1. From this figure we can see that after resampling the unavailable proposals the probabilities of the birth and death moves become equal approximately 0.17 and 0.32, i.e., the death moves are made with a probability which is significantly larger than a probability originally set equal 0.2.

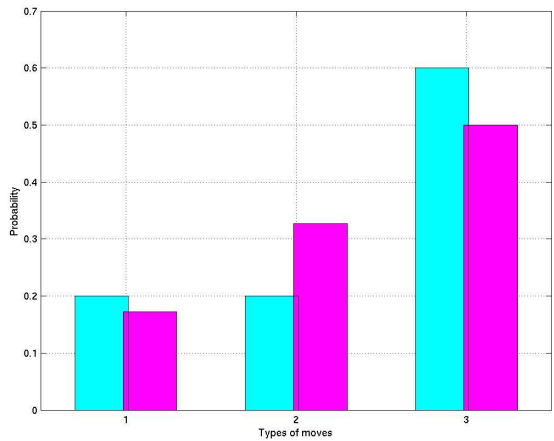

Fig. 1. The proposal probabilities for the birth, death, and change moves denoted by the first, second, and third groups, respectively. The left hand bars in each group denote the proposal probabilities set equal to 0.2, 0.2, and 0.6, respectively. The right hand bars in these groups denote the resultant probabilities with which the birth, death, and change moves are made in reality if the birth and change moves were assigned unavailable with probabilities 0.1 and 0.3, respectively.

The disproportion in the balance between the probabilities of birth and death moves is dependent on the size of DTs averaged over samples. Clearly, at the beginning of burn-in phase the disproportion is close to zero, and to the end of the burn-in phase, when the size and form of DTs are stabilized, its value becomes maximal.

The DTs grow very quickly during the first burn-in samples because an increase in log likelihood value for the birth moves is much larger than that for the others. For this reason almost every new partition of data is accepted. Once a DT has grown the *change* moves are accepted with a very small probability and, as a result, the RJ MCMC algorithm tends to get stuck at a particular DT structure instead of exploring all possible structures.

Because DTs are hierarchical structures, the changes at the nodes located at the upper levels can significantly change the location of data points at the lower levels. For this reason there is a very small probability of changing and then accepting a DT located near a root node. Therefore the RJ MCMC algorithm collects the DTs in which the splitting nodes located far from a root node were changed. These nodes typically contain small numbers of data points. Subsequently, the value of log likelihood is not changed much, and such moves are frequently accepted. As a result, the RJ MCMC algorithm cannot explore a full posterior distribution properly.

One way to extend the search space is to restrict DT sizes during a given number of the first burn-in samples as described in [5]. Indeed, under such a restriction, this strategy gives more chances of finding DTs of a smaller size which could be competitive in term of the log likelihood values with the larger DTs. The restricting strategy, however, requires setting up in an *ad hoc* manner the additional parameters such as the size of DTs and the number of the first burn-in samples. Sadly, in practice, it often happens that after the limitation period the DTs grow quickly again and this strategy does not improve the performance.

Alternatively to the above approach based on the explicit limitation of DT size, the search space can be extended by using a restarting strategy as Chipman *et al.* have suggested in [3]. Clearly, both these strategies cannot guarantee that most of DTs will be sampled from a model space region with a maximal posterior. In the next section we describe our approach based on sweeping the DTs.

### III. THE BAYESIAN AVERAGING WITH A SWEEPING STRATEGY

In this section we describe our approach to decreasing the uncertainty of classification outcomes within the Bayesian averaging over DT models. The main idea of this approach is to assign the prior probability of further splitting DT nodes to be dependent on the range of values within which the number of data points will be not less than a given number of points, $p_{min}$. Such a prior is explicit because at the current partition the range of such values is unknown.

Formally, the probability $P_s(i, j)$ of further splitting at the $i$th partition level and variable $j$ can be written as

$$P_s(i, j) = \frac{x_{max}^{(i,j)} - x_{min}^{(i,j)}}{x_{max}^{(1,j)} - x_{min}^{(1,j)}}, \qquad (13)$$

where $x_{min}^{(i,j)}$ and $x_{max}^{(i,j)}$ are the minimal and maximal values of variable $j$ at the $i$th partition level.

Observing Eq. (13), we can see that $x_{max}^{(i,j)} \leq x_{max}^{(1,j)}$ and $x_{min}^{(i,j)} \geq x_{max}^{(1,j)}$ for all the partition levels $i > 1$. On the other hand there is partition level $k$ at which the number of data points becomes less than a given number $p_{min}$. Therefore, we can conclude that the prior probability of splitting $P_s$ ranges between 0 and 1 for any variable $j$ and the partition levels $i$: $1 \leq i < k$.

From Eq. (13) it follows that for the first level of partition, probability $P_s$ is equal to 1.0 for any variable $j$. Let us now assume that the first partition split the original data set into two non-empty parts. Each of these parts contains less data points than the original data set, and consequently for the ($i = 2$)th partition either $x_{max}^{(i,j)} < x_{max}^{(1,j)}$ or $x_{min}^{(i,j)} > x_{max}^{(1,j)}$ for new splitting variable $j$. In any case, numerator in (13) decreases, and probability $P_s$ becomes less than 1.0. We can see that each new partition makes values of numerator and consequently probability (13) smaller. So the probability of further splitting nodes is dependent on the level $i$ of partitioning data set.

The above prior favors splitting the terminal nodes which contain a large number of data points. This is clearly a desired property of the RJ MCMC technique because it allows accelerating the convergence of Markov chain. As a result of using prior (13), the RJ MCMC technique of sampling DTs can explore an area of a maximal posterior in more detail.

However, prior (13) is dependent not only on the level of partition but also on the distribution of data points in the partitions. Analyzing the data set at the *i*th partition, we can see that value of probability $P_s$ is dependent on the distribution of these data. For this reason the prior (13) cannot be implemented explicitly without the estimates of the distribution of data points in each partition.

To make the birth and change moves within prior (13), the new splitting values $s_i^{\text{rule,new}}$ for the *i*th node and variable *j* are assigned as follows. For the birth and change-split moves the new value $s_i^{\text{rule,new}}$ is drawn from a uniform distribution:

$$s_i^{\text{rule,new}} \sim U(x_{\min}^{1,j}, x_{\max}^{1,j}). \quad (14)$$

The above prior is "uninformative" and used when no information on preferable values of $s_i^{\text{rule}}$ is available. As we can see, the use of a uniform distribution for drawing new rule $s_i^{\text{rule,new}}$, proposed at the level $i > 1$, can cause the partitions containing less the data points than $p_{min}$. However, within our technique such proposals can be avoided.

For the change-split moves, drawing $s_i^{\text{rule,new}}$ follows after taking new variable $s_i^{\text{var,new}}$:

$$s_i^{\text{var,new}} \sim U\{S_k\}, \quad (15)$$

where $S_k = \{1, \ldots, m\} \setminus s_i^{\text{var}}$ is the set of features excluding variable $s_i^{\text{var}}$ currently used at the *i*th node.

For the change-rule moves, the value $s_i^{\text{rule,new}}$ is drawn from a Gaussian with a given variance $\sigma_j$:

$$s_i^{\text{rule,new}} \sim N(s_i^{\text{rule}}, \sigma_j), \quad (16)$$

where $j = s_i^{\text{var}}$ is the variable used at the *i*th node.

Because DTs have hierarchical structure, the change moves (especially change-split moves) applied to the first partition levels can heavily modify the shape of the DT, and as a result, its bottom partitions can contain less the data points than $p_{min}$. As mentioned in section II, within the Bayesian DT techniques [3, 5] such moves are assigned unavailable.

Within our approach after birth or change move there arise three possible cases. In the first case, the number of data points in each new partition is larger than $p_{min}$. The second case is where the number of data points in one new partition is larger than $p_{min}$. The third case is where the number of data points in two or more new partitions is larger than $p_{min}$. These three cases are processed as follows.

For the first case, no further actions are made, and the RJ MCMC algorithm runs as usual.

For the second case, the node containing unacceptable number of data points is removed from the resultant DT. If the move was of birth type, then the RJ MCMC resamples the DT. Otherwise, the algorithm performs the death move.

For the last case, the RG MCMC algorithm resamples the DT.

As we can see, within our approach the terminal node, which after making the birth or change moves contains less than $p_{min}$ data points, is removed from the DT. Clearly, removing such unacceptable nodes turns the random search in a direction in which the RJ MCMC algorithm has more chances to find a maximum of the posterior amongst shorter DTs. As in this process the unacceptable nodes are removed, we named such a strategy *sweeping*.

After change move the resultant DT can contain more than one nodes splitting less than $p_{min}$ data points. However this can happen at the beginning of burn-in phase, when the DTs grow, and this unlikely happen, when the DTs have grown.

As an example, Fig. 2 provides the resultant probabilities estimated on 10000 moves for a case when the original probabilities of the birth, death, and change moves were set equal 0.2, 0.2, and 0.6, respectively, as assumed at the example given in section II. The probabilities of the unacceptable birth and change moves were set equal 0.07 and 0.2. These values are less than those that were set in the previous example because the DTs induced with a sweeping strategy are shorter than those induced with the standard strategy. The shorter DTs, the more data points fall at their splitting nodes, and less the probabilities $p_{bu}$ and $p_{cu}$ are. In addition, 1/10th of the unacceptable change moves was set assigned to the third option, mentioned above, for which two or more new partitions contain less than $p_{min}$ data points.

From Fig. 2 we can see that after resampling the unacceptable birth moves and reassigning the unacceptable change moves, the resultant probabilities of the birth and death moves become equal approximately 0.17 and 0.3, i.e., the values of these probabilities are very similar to those that shown in Fig. 1.

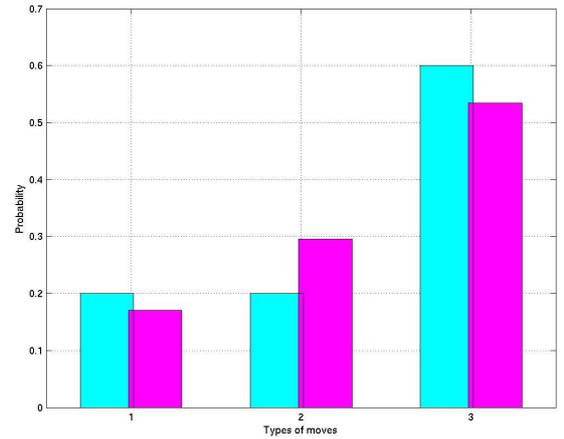

Fig. 2. The proposal probabilities for the birth, death, and change moves denoted by the first, second, and third groups, respectively. The left hand bars in each group denote the proposal probabilities set equal to 0.2, 0.2, and 0.6, respectively. The right hand bars in these groups denote the resultant probabilities with which the birth, death, and change moves are made in reality when the unacceptable birth moves are redone and the unacceptable change moves are reassigned with probabilities 0.07 and 0.2, respectively.

Next we describe the experimental results obtained with the suggested strategy of the Bayesian averaging over DTs. These results are then compared with those that have been obtained with the standard Bayesian DT technique described in [5].

## IV. EXPERIMENTAL RESULTS

For experimental evaluation of the proposal Bayesian MCMC strategy we used the following data sets. The first is an artificial exclusive OR problem (XOR3), in which the output $y = \text{sign}(x_1 x_2)$, where $x_1, x_2 \sim U(-0.5, 0.5)$ and $x_3 \sim N(0, 0.2)$ is a noise variable. Three other problems are taken from the UCI Machine Learning Repository [8], and the last is a real problem for which it is required to predict survival probability of patient after injury. Table I lists the number of classes $C$, number of variables $m$, and number of patterns $n$ for these data sets.

TABLE I
THE CHARACTERISTICS OF DATASETS

| # | Data | C | m | n |
|---|------|---|---|---|
| 1 | XOR3 | 2 | 3 | 1000 |
| 2 | Ionosphere | 2 | 33 | 351 |
| 3 | Votes | 2 | 16 | 435 |
| 4 | Wisconsin | 2 | 9 | 683 |
| 5 | Injury | 2 | 18 | 1468 |

The first, XOR3, problem is resolved with the DTs consisting of three nodes. The pruning factor $p_{min}$ was set equal to 5. The proposal probabilities for the death, birth, change-split and change-rules are set to be 0.1, 0.1, 0.1, and 0.7, respectively. The number of burn-in, post burn-in samples, and the sampling rate were 50000, 10000, and 7, respectively.

The resultant Bayesian DTs perform quite well, recognizing 100.0% of the test examples. The acceptance rate was 0.25 for burn-in and 0.25 for post burn-in. The average number of DT nodes and $2\sigma$ interval within 5 fold cross-validation were 3.8 and 0.4, respectively.

Fig. 3 depicts samples of log likelihood and numbers of DT nodes as well as the densities of DT nodes for burn-in and post burn-in phases. From the top left plot of this figure we can see that the Markov chain very quickly converges to the stationary value of log likelihood near to zero. During post burn-in the values of log likelihood slightly oscillate around zero as depicted in Fig. 3.

Table II provides the performance rates of the standard Bayesian DT strategy (BDT1) and our Bayesian DT technique with a sweeping strategy (BDT2), calculated for all the 5 data sets within 5 fold cross-validation. Both these techniques ran with the same proposal probabilities and the value $p_{min}$.

As we can see from Table II, both BDT1 and BDT2 strategies reveal the same performance on the test data. It is important to note that in these experiments the prior information, such as the preferable number of nodes and the minimal number $p_{min}$, has not been available. The number $p_{min}$ was set equal to 3 for the first four domain problems XOR3, Ionosphere, Votes, and Wisconsin, and equal to 5 for the last relatively large Injury problem.

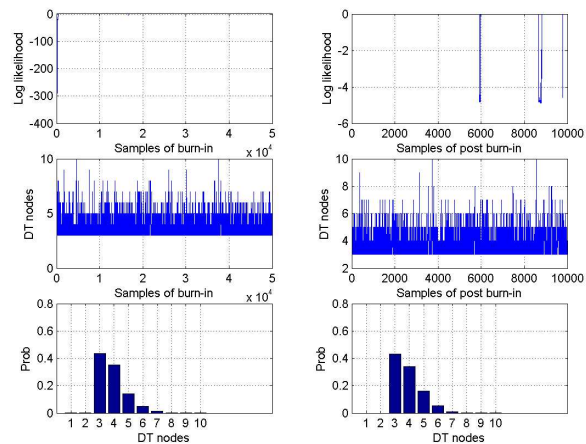

Fig. 3. Synthetic XOR data: Samples of log likelihood and DT size during burn-in (the left side) and post burn-in (the right side) phases. The bottom plots depict the distributions of DT sizes.

TABLE II
THE PERFORMANCE OF BDT1 AND BDT2

| # | Data | BDT1, % | BDT2, % |
|---|------|---------|---------|
| 1 | XOR3 | 100.0±0.0 | 100.0±0.0 |
| 2 | Ionosphere | 91.1±6.7 | 91.1±7.4 |
| 3 | Votes | 95.4±5.1 | 96.6±5.1 |
| 4 | Wisconsin | 96.2±3.8 | 96.0±3.5 |
| 5 | Injury | 84.1±3.0 | 84.6±5.2 |

The classification uncertainty of the BDT1 and BDT2 techniques is compared in terms of the sum entropy as described in [7]. The entropy $E$ is summed over class posterior probabilities $P_{ij}$, calculated for the $i$th test datum and the $j$th class as follows

$$E = -\sum_{i=1}^{t} \sum_{j=1}^{C} P_{ij} \log(P_{ij}), \qquad (17)$$

where $t$ is the number of the test examples.

Table III provides the values of entropy $E$ calculated for evaluating of classification uncertainty of the BDT1 and BDT2 on the test data within the 5 fold-cross validation.

TABLE III
THE CLASSIFICATION UNCERTAINTY IN TERM OF ENTROPY

| # | Data | BDT1 | BDT2 |
|---|------|------|------|
| 1 | XOR3 | 2.7±1.5 | **0.72**±1.5 |
| 2 | Ionosphere | 165.7±4.2 | **14.4**±5.6 |
| 3 | Votes | 48.2±4.7 | **9.7**±2.9 |
| 4 | Wisconsin | 143±2.9 | **11.4**±6.8 |
| 5 | Injury | 179.8±10.4 | **95.0**±7.6 |

From Table III we can see that in terms of the classification uncertainty the proposal strategy BDT2 is significantly superior to the BDT1.

Table IV shows the average number of nodes in DTs and $2\sigma$ intervals calculated for the BDT1 and BDT2.

TABLE IV
THE NUMBER OF NODES IN DECISION TREES

| # | Data | BDT1 | BDT2 |
|---|------|------|------|
| 1 | XOR3 | 18.5±3.9 | **3.8**±0.4 |
| 2 | Ionosphere | 16.9±1.6 | **11.9**±2.4 |
| 3 | Votes | 11.4±3.4 | **7.2**±1.2 |
| 4 | Wisconsin | 13.4±0.9 | **9.0**±1.7 |
| 5 | Injury | 43.1±5.4 | **25.0**±1.8 |

From Table IV we can see that on all the data sets the BDT2 strategy, providing the same performances, produces much shorter DTs than the BDT1 strategy. Clearly, this is a desired property of the proposed Bayesian DT technique.

## V. CONCLUSION

The use of the RJ MCMC methodology of stochastic sampling from the posterior distribution makes Bayesian DT techniques feasible. However, exploring the space of DTs parameters, existing techniques may prefer sampling DTs from the local maxima of the posterior instead of the properly representing the posterior. This affects the evaluation of the posterior distribution and, as a result, causes an increase in the classification uncertainty. This negative effect can be reduced by averaging the DTs obtained in different starts [3] or by restricting the size of DTs during burn-in phase [5].

As an alternative way of reducing the negative effect, we have suggested the Bayesian DT technique using the sweeping strategy. Within this strategy, DTs are modified after birth or change moves by removing the splitting nodes containing fewer data points than acceptable.

We have compared the performances of the Bayesian DT techniques with the standard and the sweeping strategies on a synthetic dataset as well as on some datasets from the Machine Learning Repository and real injury data. Quantitatively evaluating the uncertainty in terms of entropy, we have found that our Bayesian DT technique using the sweeping strategy is superior to the standard Bayesian DT technique. We also observe that the sweeping strategy provides much shorter DTs.

Thus we conclude that our Bayesian strategy of averaging DTs is able decreasing the classification uncertainty without affecting classification accuracy on the problems examined. Clearly this is a very desirable property for classifiers used in safety-critical systems in which classification uncertainty is of crucial importance.